%% file: main.tex
\documentclass[10pt,twocolumn,letterpaper]{article}

\pdfoutput=1

\input{Sections/packages}
\begin{document}

\title{FlashDecoder: Real-Time Latent-to-Pixel Streaming Decoder with Transformers}

\author{Minguk Kang$^{1,2}$
\hspace{8mm}
Suha Kwak$^{2}$
\vspace{2.5mm}\\
$^{1}$Pika Labs
\hspace{12mm}
$^{2}$POSTECH
}

\maketitle
\input{Sections/macros}
\input{Sections/0_abstract}
\input{Table_Figure/teaser}
\input{Sections/1_introduction}

\input{Sections/2_background}
\input{Sections/3_method}
\input{Sections/4_experiments}
\input{Sections/5_conclusion}

{
    \small
    \bibliographystyle{ieeenat_fullname}
    \bibliography{egbib}
}

\clearpage
\maketitlesupplementary

\setcounter{table}{0}
\renewcommand{\thetable}{\Alph{table}}
\renewcommand*{\theHtable}{\thetable}
\setcounter{figure}{0}
\renewcommand{\thefigure}{\Alph{figure}}
\renewcommand*{\theHfigure}{\thefigure}
\setcounter{section}{0}
\renewcommand{\thesection}{\Alph{section}}
\renewcommand*{\theHsection}{\thesection}

\input{Sections/7_appendix}

\end{document}

%% file: Sections/packages.tex
\usepackage[final]{cvpr}

\definecolor{cvprblue}{rgb}{0.21,0.49,0.74}

\usepackage{graphicx}
\usepackage{amsmath}
\usepackage{amssymb}
\usepackage{booktabs}
\usepackage[dvipsnames]{xcolor}
\usepackage{colortbl}
\usepackage{dsfont}
\usepackage{multirow}
\usepackage{makecell}
\usepackage[pagebackref,breaklinks,colorlinks,allcolors=cvprblue]{hyperref}
\usepackage{algorithm}
\usepackage[noend]{algpseudocode}

\PassOptionsToPackage{sort,comma,numbers}{natbib}
\usepackage[capitalize]{cleveref}
\crefname{section}{Sec.}{Secs.}
\Crefname{section}{Section}{Sections}
\Crefname{table}{Table}{Tables}
\crefname{table}{Tab.}{Tabs.}

\def\paperID{}
\def\confName{CVPR}
\def\confYear{2026}

%% file: Sections/macros.tex
\newcommand{\minguk}[1]{}
\newcommand{\suha}[1]{}

\definecolor{darkblue}{HTML}{202282}
\definecolor{lightgray}{HTML}{E1E1E1}
\newcommand{\newtext}[1]{#1}

\newcommand{\perm}{\boldsymbol{\pi}}

\newcommand{\myparagraph}[1]{\vspace{-5pt}\paragraph{#1}}
\newcommand{\mysubsection}[1]{\vspace{1mm}\noindent{\bf #1}}

\newcommand{\cmark}{\ding{51}}
\newcommand{\xmark}{\ding{55}}

\def\L{\mathcal{L}}

\def\vpos{\bm{v^{+}}}
\def\vneg{\bm{v^{-}}}

\def\X{\mathcal{X}}
\def\Y{\mathcal{Y}}
\def\Z{\mathcal{Z}}
\def\data{\mathcal{D}}
\def\P{\mathbb{P}}
\def\Px{\mathbb{P}_{\mathcal{X}}}
\def\Pg{\mathbb{P}_{\theta}}
\def \Lv{\mathcal{L}_{\text{models}}}
\def \kmax{K\max}

\def\xhat{{\hat{\bm x}}}
\def\yhat{{\hat{\bm y}}}
\def\zhat{{\hat{\bm z}}}

\def\xtilde{{\tilde{\x}}}
\def\ytilde{{\tilde{\y}}}
\newcommand*\wc{{\mkern 2mu\cdot\mkern 2mu}}

\def\ztilde{{\tilde{\z}}}

\def\xbar{{\bar{\x}}}
\def\ybar{{\bar{\y}}}
\def\zbar{{\bar{\z}}}

\def\xstar{{\x^{*}}}
\def\ystar{{\y^{*}}}
\def\zstar{{\z^{*}}}

\def\D{{D}}
\def\G{{G}}
\def\Re{\mathds{R}}
\newcommand{\expect}[1]{\mathbb{E}_{#1}}
\def\NCE{\ell}

\newcommand{\citeColored}[2]{{\hypersetup{citecolor=#1}\cite{#2}}}

\newcommand{\fid}{Fr\'echet Inception Distance\xspace}

\def\rvepsilon{{\mathbf{\epsilon}}}
\def\rvtheta{{\mathbf{\theta}}}
\def\rva{{\mathbf{a}}}
\def\rvb{{\mathbf{b}}}
\def\rvc{{\mathbf{c}}}
\def\rvd{{\mathbf{d}}}
\def\rve{{\mathbf{e}}}
\def\rvf{{\mathbf{f}}}
\def\rvg{{\mathbf{g}}}
\def\rvh{{\mathbf{h}}}
\def\rvu{{\mathbf{i}}}
\def\rvj{{\mathbf{j}}}
\def\rvk{{\mathbf{k}}}
\def\rvl{{\mathbf{l}}}
\def\rvm{{\mathbf{m}}}
\def\rvn{{\mathbf{n}}}
\def\rvo{{\mathbf{o}}}
\def\rvp{{\mathbf{p}}}
\def\rvq{{\mathbf{q}}}
\def\rvr{{\mathbf{r}}}
\def\rvs{{\mathbf{s}}}
\def\rvt{{\mathbf{t}}}
\def\rvu{{\mathbf{u}}}
\def\rvv{{\mathbf{v}}}
\def\rvw{{\mathbf{w}}}
\def\rvx{{\mathbf{x}}}
\def\rvy{{\mathbf{y}}}
\def\rvz{{\mathbf{z}}}

\newcommand{\seg}{\rvs_c}
\newcommand{\repr}{\rvr}
\newcommand{\repru}{\repr_{u, \sP}}
\newcommand{\uprepru}{\repr_{u, \sP}^{\uparrow}}
\newcommand{\f}{f}
\newcommand{\h}{h}

\newcommand{\figleft}{{\em (Left)}}
\newcommand{\figcenter}{{\em (Center)}}
\newcommand{\figright}{{\em (Right)}}
\newcommand{\figtop}{{\em (Top)}}
\newcommand{\figbottom}{{\em (Bottom)}}
\newcommand{\captiona}{{\em (a)}}
\newcommand{\captionb}{{\em (b)}}
\newcommand{\captionc}{{\em (c)}}
\newcommand{\captiond}{{\em (d)}}

\newcommand{\newterm}[1]{{\bf #1}}

\newcommand{\reffig}[1]{Figure~\ref{fig:#1}}
\newcommand{\refsec}[1]{Section~\ref{sec:#1}}
\newcommand{\refapp}[1]{Appendix~\ref{sec:#1}}
\newcommand{\reftbl}[1]{Table~\ref{tbl:#1}}
\newcommand{\refalg}[1]{Algorithm~\ref{alg:#1}}
\newcommand{\refline}[1]{Line~\ref{line:#1}}
\newcommand{\shortrefsec}[1]{\S~\ref{sec:#1}}
\newcommand{\refeqshort}[1]{(\ref{eq:#1})}
\newcommand{\shortrefeq}[1]{\ref{eq:#1}}
\newcommand{\lblfig}[1]{\label{fig:#1}}
\newcommand{\lblsec}[1]{\label{sec:#1}}
\newcommand{\lbleq}[1]{\label{eq:#1}}
\newcommand{\lbltbl}[1]{\label{tbl:#1}}
\newcommand{\lblalg}[1]{\label{alg:#1}}
\newcommand{\lblline}[1]{\label{line:#1}}
\newcommand{\ignorethis}[1]{}
\newcommand{\revision}[1]{\color{black}#1\color{black}}
\newcommand{\myitem}{\vspace{-5pt}\item}

\def\ceil#1{\lceil #1 \rceil}
\def\floor#1{\lfloor #1 \rfloor}
\def\1{\bm{1}}
\newcommand{\train}{\mathcal{D}}
\newcommand{\valid}{\mathcal{D_{\mathrm{valid}}}}
\newcommand{\test}{\mathcal{D_{\mathrm{test}}}}

\def\eps{{\epsilon}}

\newcommand{\image}{{\rvx}}
\newcommand{\latent}{{\rvz}}
\newcommand{\images}{{\mathcal{X}}}
\newcommand{\imagedist}{{p_{\text{data}}(\image)}}
\newcommand{\latentdist}{{p(\latent)}}
\newcommand{\imageD}{{D_X}}
\newcommand{\Fnet}{{F}}
\newcommand{\sketch}{{\rvy}}
\newcommand{\sketches}{{\mathcal{Y}}}
\newcommand{\sketchdist}{{p_{\text{data}}(\sketch)}}
\newcommand{\sketchD}{{D_Y}}
\newcommand{\modelold}{{G(\rvz; \theta)}}
\newcommand{\modelnew}{{G(\rvz; \theta')}}
\newcommand{\losssketch}{{\mathcal{L}_{\text{sketch}}}}
\newcommand{\lossimage}{{\mathcal{L}_{\text{image}}}}
\newcommand{\lossweight}{{\mathcal{L}_{\text{weight}}}}

\newcommand{\method}{{GAN Sketching}}

\newcommand{\pdata}{{D}}
\newcommand{\ptrain}{\hat{p}_{\rm{data}}}
\newcommand{\Ptrain}{\hat{P}_{\rm{data}}}
\newcommand{\pmodel}{p_{\rm{model}}}
\newcommand{\Pmodel}{P_{\rm{model}}}
\newcommand{\ptildemodel}{\tilde{p}_{\rm{model}}}
\newcommand{\pencode}{p_{\rm{encoder}}}
\newcommand{\pdecode}{p_{\rm{decoder}}}
\newcommand{\precons}{p_{\rm{reconstruct}}}

\newcommand{\laplace}{\mathrm{Laplace}} %

\newcommand{\Ls}{\mathcal{L}}
\newcommand{\R}{\mathbb{R}}
\newcommand{\emp}{\tilde{p}}
\newcommand{\lr}{\alpha}
\newcommand{\reg}{\lambda}
\newcommand{\rect}{\mathrm{rectifier}}
\newcommand{\softmax}{\mathrm{softmax}}
\newcommand{\sigmoid}{\sigma}
\newcommand{\softplus}{\zeta}
\newcommand{\Var}{\mathrm{Var}}
\newcommand{\standarderror}{\mathrm{SE}}
\newcommand{\Cov}{\mathrm{Cov}}
\newcommand{\normlzero}{L^0}
\newcommand{\normlone}{L^1}
\newcommand{\normltwo}{L^2}
\newcommand{\normlp}{L^p}
\newcommand{\normmax}{L^\infty}

\newcommand{\parents}{Pa} %

\newcommand{\xpar}[1]{\noindent\textbf{#1}\ \ }
\newcommand{\vpar}[1]{\vspace{3mm}\noindent\textbf{#1}\ \ }

\newcommand{\shapenet}{ShapeNet\xspace}
\newcommand{\pascal}{PASCAL 3D+\xspace}

\newcommand{\degree}{\ensuremath{^\circ}\xspace}
\newcommand{\ignore}[1]{}

\newcommand{\fcseven}{$\mbox{fc}_7$}

\renewcommand*{\thefootnote}{\arabic{footnote}}

\def\naive{na\"{\i}ve\xspace}
\def\Naive{Na\"{\i}ve\xspace}

\makeatletter
\DeclareRobustCommand\onedot{\futurelet\@let@token\@onedot}
\def\@onedot{\ifx\@let@token.\else.\null\fi\xspace}
\def\eg{e.g\onedot,\xspace} 
\def\Eg{E.g\onedot,}
\def\ie{i.e\onedot,\xspace} 
\def\Ie{\emph{I.e}\onedot,}
\def\cf{\emph{c.f}\onedot} \def\Cf{\emph{C.f}\onedot}
\def\etc{\emph{etc}\onedot} \def\vs{\emph{vs}\onedot}
\def\wrt{w.r.t\onedot} \def\dof{d.o.f\onedot}
\def\etal{\emph{et al}\onedot}
\makeatother

\newcommand*{\img}[1]{%
    \raisebox{-.25\baselineskip}{%
        \includegraphics[
        height=\baselineskip,
        width=\baselineskip,
        keepaspectratio,
        ]{#1}%
    }%
}

\fboxsep=0mm%
\fboxrule=2pt%
\newcommand*{\myoverpic}[2]{
  \begin{overpic}[width=.18\linewidth]{#1}
     \put(60,63){\fbox{\includegraphics[width=.06\linewidth]{#2}}}  
  \end{overpic}
}

\newcommand{\OursAcronym}{Diff2pix}

\newcommand{\figwidth}{1.0\linewidth}

%% file: Sections/0_abstract.tex
\begin{abstract}
Real-time video generation demands fast decoding as much as fast denoising, yet current latent video diffusion models rely on 3D convolutional decoders that are slow and memory-intensive at high resolutions or for long video. We introduce \textbf{FlashDecoder}, a fast, memory-efficient pure-Transformer video decoder that decodes latents to pixels frame by frame. At each step, the current frame attends only to a fixed-size window of past frames through a rolling KV cache. The fixed temporal window keeps decoding fast and memory bounded regardless of video length, enabling constant-latency streaming. Because frames are processed sequentially, temporal causality is enforced without explicit attention masks, enabling training at resolutions up to 1080p and matching the reconstruction quality of convolutional decoders. On the Wan2.1 and Wan2.2 latent spaces, FlashDecoder matches each convolutional decoder in reconstruction quality (e.g., 41.55 vs.\ 41.49\,dB PSNR at 1080p) while decoding $3.6{\times}$--$4.7{\times}$ faster with up to $11{\times}$ less memory on a single H100 GPU. With architecture-aware inference optimizations, the speedup widens to $12{\times}$.
\end{abstract} 

%% file: Table_Figure/teaser.tex
\begin{figure}[t!]
  \centering
  \includegraphics[width=0.95\linewidth]{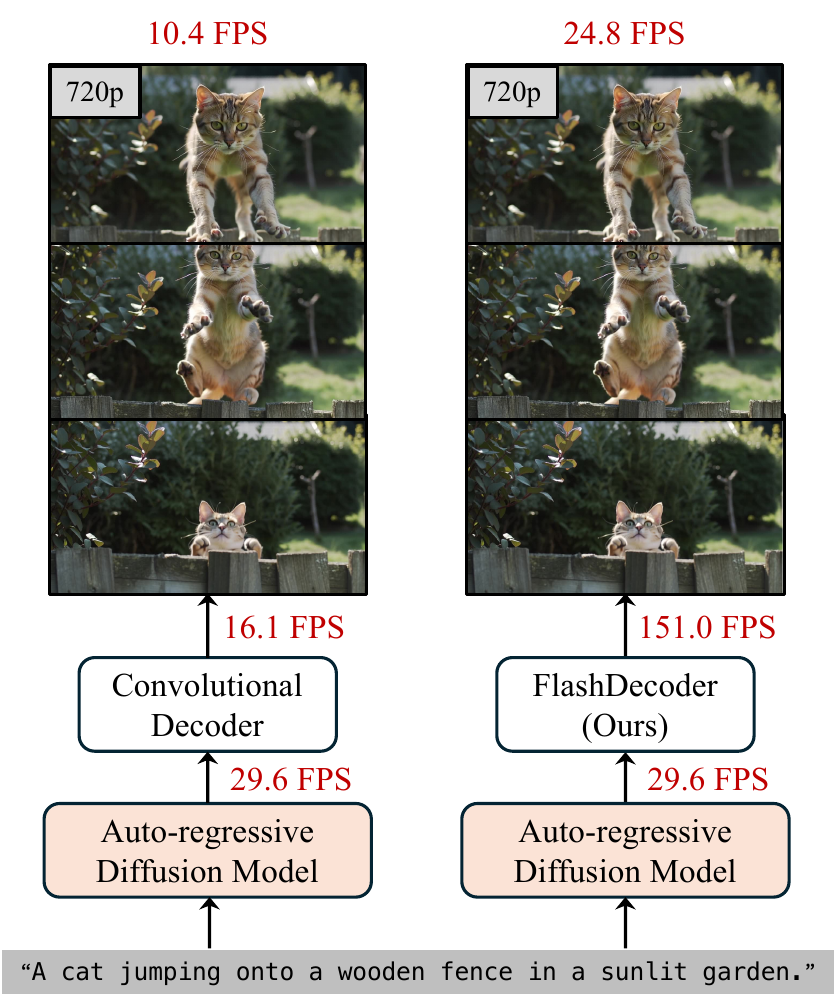}
    \caption{\textbf{VAE decoding is a major bottleneck for real-time video generation.} Measured with our MotionStream~\cite{shin2025motionstream} implementation at 720p. The Wan2.2~\cite{wan2025} decoder consumes 64.6\% of total inference time, limiting generation to 10.4 FPS. FlashDecoder reduces this share to 16.4\%, more than doubling end-to-end throughput to 24.8 FPS.}
  \label{fig:teaser}
\end{figure}

%% file: Sections/1_introduction.tex
\section{Introduction}
Latent diffusion models~\cite{rombach2022high} have become the dominant framework for image~\cite{ramesh2022hierarchical, esser2024scaling, batifol2025flux} and video generation~\cite{videoworldsimulators2024, gao2025seedance, wan2025, veo2024, ltx2}. An encoder~\cite{hinton2006reducing, kingma2013auto} compresses pixels into a low-dimensional latent space where diffusion models~\cite{ho2020denoising, song2020score, nichol2021improved} generate content, and a decoder then reconstructs the final output pixels. By operating in latent space rather than pixel space, this design reduces computational costs by orders of magnitude, enabling scaling to billions of parameters with Diffusion Transformers (DiT)~\cite{peebles2023scalable, esser2024scaling}. However, most progress has focused on accelerating the generation stage, while the decoder that maps latents back to pixels has remained largely convolutional and comparatively underexplored.

Originally, the main inference bottleneck was iterative denoising. Since then, more efficient architectures~\cite{yao2024fasterdit, zhu2025dig, gao2025seedance, xie2024sana}, higher-compression VAEs~\cite{chendeep, hacohen2024ltx, agarwal2025cosmos, yao2025reconstruction, wan2025}, and few-step distillation~\cite{salimans2022progressive, meng2023distillation, Song2023ConsistencyM, kim2023consistency, kang2024distilling, sauer2024fast, yin2024improved, chen2025sana, pmlr-v267-lin25m} have largely removed this bottleneck. But real-time streaming also requires causal frame-by-frame generation. Recent causal video generators~\cite{yin2025causvid, huang2025self, lin2025autoregressive, krea2025realtime14b, shin2025motionstream, liu2025rolling} are closing in on interactive frame rates. With latent generation nearing real-time speed, the bottleneck has shifted to the decoder: VAE decoding consumes over 64.6\% of total inference time with the Wan2.2 decoder~\cite{wan2025} at 720p (Figure~\ref{fig:teaser}).

Existing video decoders are predominantly 3D causal convolutional networks~\cite{lin2024open, yang2025cogvideox, kong2024hunyuanvideo, gao2025seedance, wan2025, zhang2025waver}. They reconstruct well but are slow and memory-intensive, making real-time streaming impractical. High-resolution decoding further requires spatial-temporal tiling, which multiplies decoder evaluations and latency. Transformer-based~\cite{vaswani2017attention} decoders have been explored as alternatives, but face a trade-off between streaming and quality. Causal variants~\cite{villegas2023phenaki, wang2024omnitokenizer} require explicit causal masks that prevent efficient FlashAttention~\cite{dao2022flashattention} usage at high resolutions, limiting reconstruction fidelity. Bidirectional variants~\cite{teng2025magi, atoken} achieve better quality by attending across all frames, but cannot stream because each frame requires access to future frames.

The limitations of both convolutional and Transformer decoders suggest four desired properties for a real-time video decoder: (1) frame-by-frame decoding without padding or blending, (2) reconstruction quality competitive with convolutional decoders, (3) fast inference with consistent per-frame latency and bounded memory, and (4) high-resolution and long-duration decoding without spatial-temporal tiling. Transformers can in principle provide all four: sequential processing enables frame-by-frame decoding, self-attention captures spatial-temporal dependencies for high-quality reconstruction, and windowed attention bounds memory and computation per frame, ensuring consistent latency without tiling. However, no existing Transformer-based decoder satisfies all four properties.

We introduce \textbf{FlashDecoder}, a pure-Transformer latent-to-pixel decoder that satisfies all four properties. FlashDecoder processes one latent frame at a time with a fixed-size rolling KV cache. The fixed temporal window keeps per-frame computation and memory bounded regardless of video length, enabling constant-latency streaming without spatial-temporal tiling. Prior causal Transformer decoders train with explicit causal masks but switch to KV caching at inference. FlashDecoder instead uses the same temporal-order processing for both training and inference, removing the need for attention masks entirely. This makes high-resolution training feasible, matching the reconstruction quality of convolutional decoders. Evaluated on the Wan2.1 and Wan2.2~\cite{wan2025} latent spaces, FlashDecoder matches each convolutional decoder in reconstruction quality (e.g., 41.55 vs.\ 41.49 PSNR at 1080p on Wan2.2) while providing $3.6{\times}$--$4.7{\times}$ higher throughput and up to $11{\times}$ lower memory on a single H100 GPU. With architecture-aware inference optimizations, the throughput gap widens to up to $12{\times}$.

\input{Table_Figure/visual_comparison}

%% file: Table_Figure/visual_comparison.tex
\begin{figure*}[t!]
  \centering
  \includegraphics[width=0.99\linewidth]{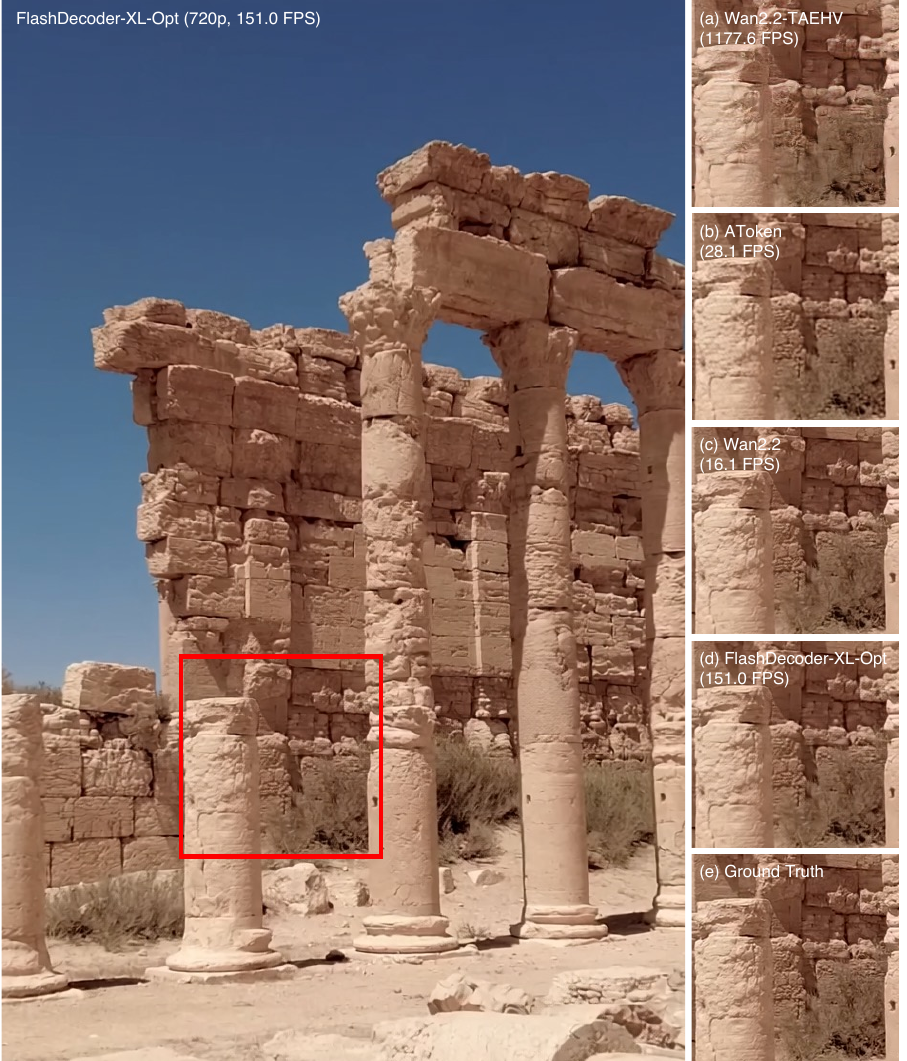}
\caption{\textbf{Qualitative comparison of 720p reconstruction results.}
We compare reconstructed frames from video decoders with 4$\times$ temporal and 16$\times$ spatial compression: (a) Wan2.2-TAEHV~\cite{BoerBohan2025TAEHV}, (b) AToken~\cite{atoken}, (c) Wan2.2~\cite{wan2025}, (d) our FlashDecoder-XL-Opt, and (e) ground truth.
(a) fails to synthesize fine details such as wall textures, while (b) produces blurry reconstructions. (c) and (d) yield visually comparable outputs, yet (d) achieves over $9{\times}$ higher throughput (151.0 vs.\ 16.1 FPS). Additional comparisons are provided in the supplementary material.}
  \label{fig:visual_comparison}
\end{figure*}

%% file: Sections/2_background.tex
\section{Related Work}
\subsection{Autoencoders for Visual Generation}
Latent diffusion models~\cite{rombach2022high} rely on learned autoencoders to compress inputs into a low-dimensional latent space. For images, autoencoders have evolved from Stable Diffusion VAE~\cite{rombach2022high} to more efficient designs such as FLUX VAE~\cite{batifol2025flux} and DC-AE~\cite{chendeep}, improving compression ratios and reconstruction quality, with most designs primarily based on convolutions. For video, these spatial autoencoders have been extended to spatiotemporal variants that compress both spatial and temporal dimensions~\cite{videoworldsimulators2024, opensora, lin2024open, yang2025cogvideox, genmo2024mochi, agarwal2025cosmos, teng2025magi, ltx2, kong2024hunyuanvideo, wan2025, gao2025seedance, zhang2025waver}. Dedicated video VAE works such as MagViT-v2~\cite{yu2024language}, CV-VAE~\cite{zhao2024cv}, WF-VAE~\cite{li2025wf}, VidTok~\cite{tang2024vidtok}, and VideoVAE+~\cite{xing2025videovae} have further advanced reconstruction fidelity through improved architectures, loss formulations, and temporal modeling. Despite this progress, spatiotemporal decoding remains computationally expensive, particularly at high resolutions. Lightweight alternatives such as TurboVAED~\cite{zou2025turbo} and TAEHV~\cite{BoerBohan2025TAEHV} trade fidelity for speed, and all of the above video decoders remain convolutional.

A separate line of work has begun moving away from convolutional decoders entirely. Representation Autoencoders such as RAE~\cite{zheng2025diffusion, tong2026scaling} and LV-RAE~\cite{liu2026improving} pair frozen pretrained encoders (e.g., DINOv2~\cite{oquab2023dinov2} and SigLIP 2~\cite{tschannen2025siglip}) with Transformer decoders, achieving faster convergence and stronger generation quality than convolutional VAEs for image generation. This trend has yet to reach video, where a streaming-capable Transformer decoder would be needed.

\subsection{Transformer-based Video Decoders}
For video, Transformer-based decoders have been explored across discrete tokenization models~\cite{yu2022vectorquantized, villegas2023phenaki}, hybrid discrete-continuous VAEs~\cite{wang2024omnitokenizer, atoken}, and fully continuous latent VAEs~\cite{teng2025magi, pmlr-v267-hansen-estruch25a}. These designs face a fundamental trade-off in how they handle temporal attention. Causal attention, as in OmniTokenizer~\cite{wang2024omnitokenizer}, enables streaming through KV caching but requires explicit causal masks during training, making memory-efficient high-resolution training difficult and limiting reconstruction quality. Per-frame latency also grows as temporal context accumulates. Bidirectional attention, as in AToken~\cite{atoken} and the MAGI-1 VAE decoder~\cite{teng2025magi}, improves quality by attending across all frames but cannot support streaming decoding due to global temporal dependencies. FlashDecoder takes the causal approach but removes its key limitations: causality is enforced by processing order rather than masks, and a fixed-size window bounds memory regardless of video length, unifying training and inference under the same streaming mechanism.

%% file: Sections/3_method.tex
\section{Method}

\input{Table_Figure/main_figure}

Figure~\ref{fig:main_figure} illustrates the overall pipeline. We first describe the latent diffusion framework (Sec.~\ref{sec:prelim}), then detail the Transformer backbone (Sec.~\ref{sec:arch}), the streaming mechanism with rolling KV cache (Sec.~\ref{sec:sequential}), the temporal-first upsampling strategy (Sec.~\ref{sec:upsample}), and training objectives (Sec.~\ref{sec:training}).

\subsection{Preliminaries: Latent Diffusion Model}
\label{sec:prelim}
FlashDecoder operates within the Latent Diffusion Model (LDM) framework~\cite{rombach2022high}. A pretrained encoder $\mathcal{E}$ maps a video $\mathbf{x}\in\mathbb{R}^{B\times C\times T\times H\times W}$ to a latent tensor $\mathbf{z}=\mathcal{E}(\mathbf{x})\in\mathbb{R}^{B\times C'\times T'\times H'\times W'}$, where $(C', T', H', W')$ are the compressed channel, temporal, and spatial dimensions. The diffusion process operates on $\mathbf{z}$, and a decoder $\mathcal{D}$ reconstructs pixels via $\hat{\mathbf{x}}=\mathcal{D}(\mathbf{z})$. FlashDecoder is agnostic to the choice of encoder; we train decoders on both the Wan2.1 and Wan2.2~\cite{wan2025} latent spaces and use Wan2.2 as the primary example throughout.

\subsection{Transformer Backbone Design}
\label{sec:arch}

\paragraph{Base Architecture.}
The decoder is a Transformer~\cite{vaswani2017attention} with stacked self-attention and feedforward layers. We use Grouped-Query Attention (GQA)~\cite{ainslie2023gqa}, which shares key-value heads across query groups to reduce KV cache memory during streaming. For stable training, we apply RMSNorm~\cite{zhang2019root} before each sublayer and normalize keys and values within attention (KV-norm)~\cite{team2024gemma}. Spatiotemporal position is encoded with 3D Rotary Positional Embeddings (3D-RoPE)~\cite{su2024roformer}, applied separately to the temporal ($T'$) and spatial ($H' \times W'$) dimensions.

\paragraph{Latent Projection.}
Each latent frame $\mathbf{z}_t$ is flattened into $L_{\text{frm}} = H'W'$ spatial tokens in raster order, and a linear layer maps the latent channels $C'$ to model dimension $D$:
\begin{equation}
\mathbf{P}=\mathrm{Linear}_{C'\rightarrow D}(\mathbf{z})\in\mathbb{R}^{B\times L\times D},
\label{eq:projection}
\end{equation}
where $L = T' \cdot L_{\text{frm}}$ is the total sequence length.

\subsection{Streaming with Rolling KV Cache}
\label{sec:sequential}
FlashDecoder processes video latents one frame at a time while maintaining a sliding-window KV cache of fixed size $W_{\text{frm}}$ frames. We set $W_{\text{frm}}{=}2$ throughout, so each latent frame attends only to itself and the immediately preceding latent frame; the effect of this choice is ablated in Sec.~\ref{sec:wfrm}. Because future frames have not yet been fed to the model, temporal causality is enforced by construction without explicit attention masks.

\paragraph{Frame-by-Frame Processing.}
Given $T'$ latent frames $\{\mathbf{z}_0, \ldots, \mathbf{z}_{T'-1}\}$, we process them sequentially. At each timestep $t$: \textbf{(1)}~the latent frame $\mathbf{z}_t$ is projected into $L_{\text{frm}}$ tokens via Eq.~\eqref{eq:projection}; \textbf{(2)}~new key-value pairs $(\mathbf{K}_{t}^{\text{new}}, \mathbf{V}_{t}^{\text{new}})$ are computed with 3D-RoPE at temporal offset $t \cdot L_{\text{frm}}$, appended to the cache, and the oldest frame is evicted if the cache exceeds $W_{\text{frm}}$ frames; \textbf{(3)}~current queries $\mathbf{Q}_{t}^{\text{new}}$ attend to the full cache. The resulting cache shape is:
\begin{equation}
\mathbf{K}_{t},\mathbf{V}_{t} \in \mathbb{R}^{B\times G\times (W_{\text{frm}}L_{\text{frm}})\times D_h},
\label{eq:kv_cache}
\end{equation}
where $G$ is the number of KV groups in GQA and $D_h$ is the head dimension.

\paragraph{Attention Pattern.}
The resulting pattern is a sliding window: within each frame, tokens attend to all $L_{\text{frm}}$ spatial positions (bidirectional); along the temporal axis, attention is restricted to the most recent $W_{\text{frm}}$ frames (causal).

\paragraph{Training--Inference Consistency.}
A distinctive property of FlashDecoder is that training and inference follow an identical streaming protocol: the model never sees more than $W_{\text{frm}}$ frames at once during either phase. Conventional approaches load all $T'$ frames into a single forward pass, which requires a full-sequence causal mask that FlexAttention~\cite{li2024flexattention} must materialize, causing out-of-memory failures at 480p, 720p, and 1080p on an H100 80\,GB GPU. FlashDecoder instead performs $T'$ sequential forward passes, each attending to at most $W_{\text{frm}} \cdot L_{\text{frm}}$ tokens with standard FlashAttention~\cite{dao2022flashattention} at per-step memory cost $O(W_{\text{frm}} \cdot L_{\text{frm}})$.
\label{sec:efficient_training}

\paragraph{Complexity Analysis.}
\label{sec:streaming}
The attention cost per latent frame is $\mathcal{O}(N W_{\text{frm}} L_{\text{frm}}^2 D_h)$, linear in the temporal window $W_{\text{frm}}$ and head count $N$, but quadratic in spatial tokens $L_{\text{frm}}$. The KV cache memory per layer is $\mathcal{O}(B\,G\,W_{\text{frm}}\,L_{\text{frm}}\,D_h)$, benefiting from GQA's reduced group count ($G \ll N$). After temporal upsampling (Sec.~\ref{sec:upsample}), refinement layers operate on $r_{\text{t}} \cdot L_{\text{frm}}$ tokens per frame with cache capacity $r_{\text{t}} W_{\text{frm}} L_{\text{frm}}$, preserving the same scaling properties.

\subsection{Temporal-First Upsampling Strategy}
\label{sec:upsample}
Progressive spatio-temporal upsampling is computationally prohibitive for Transformers. Spatial upsampling by factor $r_{\text{s}}$ increases tokens per frame by $r_{\text{s}}^2$, resulting in $O(r_{\text{s}}^4)$ attention cost ($65{,}536{\times}$ for $r_{\text{s}}{=}16$). Temporal upsampling by factor $r_{\text{t}}$ incurs only $O(r_{\text{t}}^2)$ cost ($16{\times}$ for $r_{\text{t}}{=}4$), a $4{,}096{\times}$ difference. We therefore adopt a temporal-first strategy: temporal upsampling via Transformer layers, followed by spatial upsampling via MLP and PixelShuffle~\cite{shi2016real}.

\paragraph{Step 1: Temporal Upsampling.} Starting from the backbone output $\mathbf{Y}\!\in\!\mathbb{R}^{B\times L\times D}$ (where $L = T' H' W'$), a linear layer expands channels by temporal factor $r_{\text{t}}$:
\begin{equation}
\mathbf{P}^{\text{temp}} = \mathrm{Linear}_{D \rightarrow D \cdot r_{\text{t}}}(\mathbf{Y}) \in \mathbb{R}^{B\times L\times (D \cdot r_{\text{t}})}.
\label{eq:temporal_upsample}
\end{equation}
Reinterpreting expanded channels as new temporal indices yields $\mathbf{P}^{\text{full}}\in\mathbb{R}^{B\times (T' r_{\text{t}} H' W')\times D}$.

\paragraph{Step 2: Temporal Refinement.} Two Transformer blocks process $\mathbf{P}^{\text{full}}$ using the same streaming mechanism from Sec.~\ref{sec:sequential}, with expanded window size $W_{\text{frm}}^{\text{full}} = r_{\text{t}}\cdot W_{\text{frm}}$ to preserve effective temporal context.

\paragraph{Step 3: Spatial Upsampling.} A 2-layer MLP projects features from $D$ to $C\cdot r_{\text{s}}^{2}$ channels, followed by PixelShuffle~\cite{shi2016real} to produce the final output $\hat{\mathbf{x}}\in\!\mathbb{R}^{B\times C\times (T' r_{\text{t}})\times (H'r_{\text{s}})\times (W'r_{\text{s}})}$.

\subsection{Training Objectives}
\label{sec:training}
The decoder $\mathcal{D}$ is trained with a combination of pixel-wise, perceptual, and adversarial losses:
\begin{equation}
\mathcal{L}_{\text{total}} = \lambda_{\text{L1}} \mathcal{L}_{\text{L1}} + \lambda_{\text{LPIPS}} \mathcal{L}_{\text{LPIPS}} + \lambda_{\text{adv}} \mathcal{L}_{\text{adv}},
\label{eq:loss}
\end{equation}
where $\mathcal{L}_{\text{L1}}$ is the pixel-wise L1 loss between the reconstructed video $\hat{\mathbf{x}}$ and ground truth $\mathbf{x}$, $\mathcal{L}_{\text{LPIPS}}$~\cite{zhang2018unreasonable} measures perceptual similarity in a pretrained feature space, and $\mathcal{L}_{\text{adv}}$~\cite{Goodfellow2014GAN} is computed using a 3D patch-based discriminator~\cite{isola2017image, esser2021taming}. L1 ensures pixel-level fidelity, LPIPS encourages perceptually realistic outputs, and the adversarial term produces sharp high-frequency details. Hyperparameter values are provided in the supplementary material. Algorithm~\ref{alg:decoder_stream} summarizes the full streaming pipeline.

\algrenewcommand\algorithmicrequire{\textbf{Require:}}
\newcommand{\Wfrm}{W_{\text{frm}}}
\newcommand{\Lfrm}{L_{\text{frm}}}
\begin{algorithm}[t]
\caption{Streaming Video Decoding from Latent}
\label{alg:decoder_stream}
\begin{algorithmic}[1]
\Require Latent $\mathbf{z} \in \mathbb{R}^{B \times C' \times T' \times H' \times W'}$ for any $T'$
\Require Window size $\Wfrm$
\Require Tokens per frame $\Lfrm = H'W'$
\Require Upsampling factors $r_{\text{t}}$ (temporal), $r_{\text{s}}$ (spatial)
\Require Backbone Transformer $\mathcal{B}$
\Require Temporal Refinement Transformer $\mathcal{R}$
\State Initialize $(\mathbf{K}^{\mathcal{B}}, \mathbf{V}^{\mathcal{B}}) \leftarrow \varnothing$, $(\mathbf{K}^{\mathcal{R}}, \mathbf{V}^{\mathcal{R}}) \leftarrow \varnothing$
\State Initialize output $\hat{\mathbf{x}} \leftarrow [\ ]$
\For{$t = 0, \ldots, T'-1$}
    \State $\mathbf{P}_t \leftarrow \text{Project}(\mathbf{z}[:,:,t:t+1,:,:])$
    \State \Comment{Backbone processing with RoPE offset $t \cdot \Lfrm$}
    \State $\mathbf{Y}_t, (\mathbf{K}_t^{\mathcal{B}}, \mathbf{V}_t^{\mathcal{B}}) \leftarrow \mathcal{B}(\mathbf{P}_t, \mathbf{K}^{\mathcal{B}}, \mathbf{V}^{\mathcal{B}}, t)$
    \State $\mathbf{K}^{\mathcal{B}} \leftarrow \text{concat}(\text{tail}_{(\Wfrm-1)\Lfrm}(\mathbf{K}^{\mathcal{B}}), \mathbf{K}_t^{\mathcal{B}})$
    \State $\mathbf{V}^{\mathcal{B}} \leftarrow \text{concat}(\text{tail}_{(\Wfrm-1)\Lfrm}(\mathbf{V}^{\mathcal{B}}), \mathbf{V}_t^{\mathcal{B}})$
    \State \Comment{Temporal upsampling: channel $\to$ temporal axis}
    \State $\mathbf{P}_t^{\text{temp}} \leftarrow \text{Linear}_{D \to r_{\text{t}} D}(\mathbf{Y}_t)$
    \State $\mathbf{P}_t^{\text{full}} \leftarrow \text{Reshape}(\mathbf{P}_t^{\text{temp}})$ \Comment{Yields $r_{\text{t}} \cdot \Lfrm$ tokens}
    \State \Comment{Temporal refinement with window $r_{\text{t}} \Wfrm$}
    \State \Comment{RoPE offset $t \cdot r_{\text{t}} \cdot \Lfrm$}
    \State $\mathbf{Y}_t^{\text{full}}, (\mathbf{K}_t^{\mathcal{R}}, \mathbf{V}_t^{\mathcal{R}}) \leftarrow \mathcal{R}(\mathbf{P}_t^{\text{full}}, \mathbf{K}^{\mathcal{R}}, \mathbf{V}^{\mathcal{R}}, t)$
    \State $\mathbf{K}^{\mathcal{R}} \leftarrow \text{concat}(\text{tail}_{r_{\text{t}}(\Wfrm-1)\Lfrm}(\mathbf{K}^{\mathcal{R}}), \mathbf{K}_t^{\mathcal{R}})$
    \State $\mathbf{V}^{\mathcal{R}} \leftarrow \text{concat}(\text{tail}_{r_{\text{t}}(\Wfrm-1)\Lfrm}(\mathbf{V}^{\mathcal{R}}), \mathbf{V}_t^{\mathcal{R}})$
    \State \Comment{Spatial upsampling: MLP + PixelShuffle}
    \State $\mathbf{F}_t \leftarrow \text{MLP}_{D \to C r_{\text{s}}^2}(\mathbf{Y}_t^{\text{full}})$
    \State $\hat{\mathbf{x}}_t \leftarrow \text{PixelShuffle}(\mathbf{F}_t, r_{\text{s}})$
    \State Append $\hat{\mathbf{x}}_t$ to $\hat{\mathbf{x}}$
\EndFor
\State \Return $\hat{\mathbf{x}}$
\end{algorithmic}
\end{algorithm}

%% file: Table_Figure/main_figure.tex
\begin{figure*}[t]
  \centering
  \includegraphics[width=1.0\linewidth]{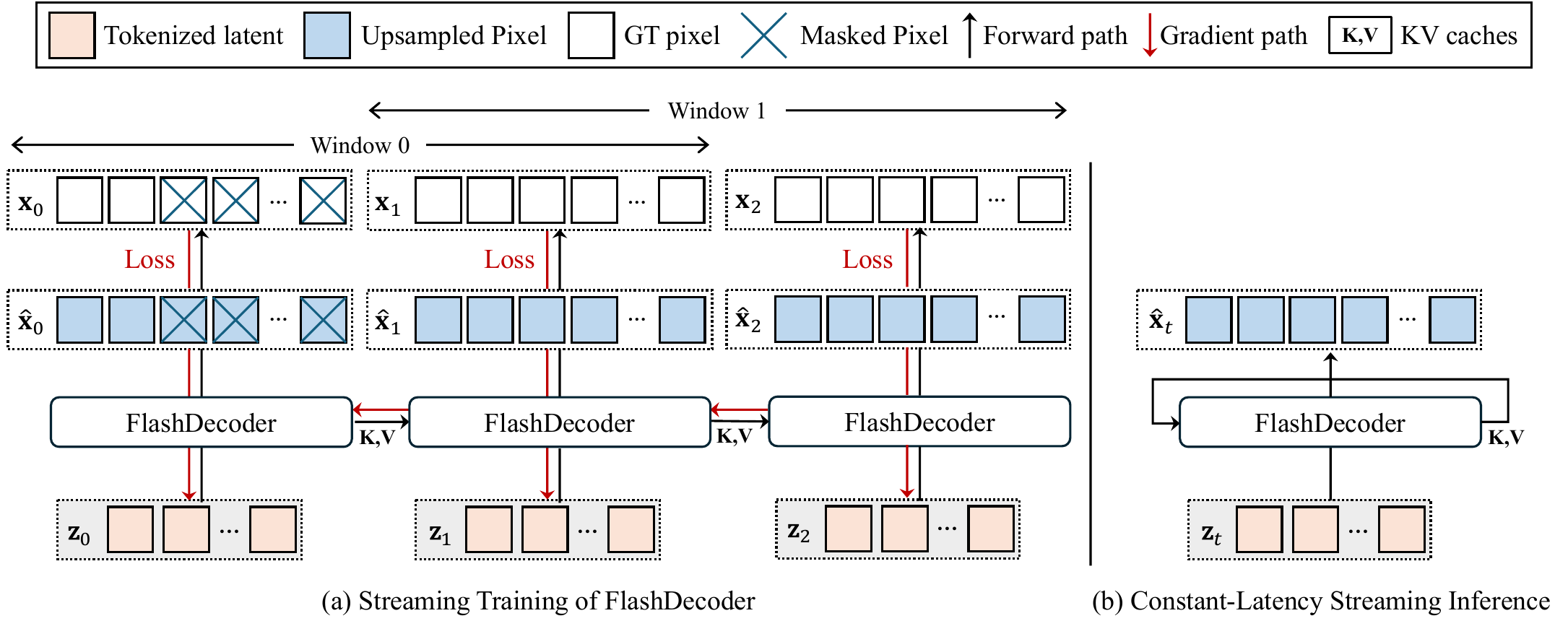}
\caption{\textbf{FlashDecoder pipeline.} FlashDecoder is a pure-Transformer decoder that converts video latents to pixels in a frame-by-frame manner. Each latent frame $\mathbf{z}_t$ is linearly projected, processed by a Transformer backbone with a fixed-size rolling KV cache that stores the most recent $W_{\text{frm}}$ frames (temporal window size), temporally upsampled by factor $r_{\text{t}}$ via channel expansion and refinement layers, and spatially upsampled via MLP and PixelShuffle. Shown here with $W_{\text{frm}}{=}2$ and $r_{\text{t}}{=}4$, streaming 3 latent frames into 9 output frames.
\textbf{(a)} Attention pattern: each frame's $L_{\text{frm}}{=}H'{\times}W'$ spatial tokens attend bidirectionally to each other and causally to the previous $W_{\text{frm}}{-}1$ frames. $\mathbf{z}_0$ is decoded alone; $\mathbf{z}_1$ attends to $[\mathbf{z}_0, \mathbf{z}_1]$; $\mathbf{z}_2$ attends to $[\mathbf{z}_1, \mathbf{z}_2]$ after evicting $\mathbf{z}_0$.
\textbf{(b)} Streaming inference: each incoming latent frame $\mathbf{z}_t$ is projected and processed through the Transformer with the rolling KV cache, then upsampled to pixel frames. The bounded cache ensures constant per-frame latency and memory regardless of video length.}
  \label{fig:main_figure}
\end{figure*}

%% file: Sections/4_experiments.tex
\section{Experiments}
Unless stated otherwise, all experiments use FlashDecoder trained on the Wan2.2 latent space (4$\times$16$\times$16). All training is conducted on a single node with 8 NVIDIA H100 GPUs.

\input{Table_Figure/ablation}
\input{Table_Figure/benchmark_video}

\subsection{Datasets and Evaluation Protocol}

\paragraph{Training Data.}
FlashDecoder is trained on joint image--video data in three stages. We use DataComp-small~\cite{gadre2023datacomp} (12.8M image--text pairs) and video data from Kinetics-600~\cite{carreira2018short} and an internal collection, with a 2:8 image-to-video sampling ratio. Stage~1 trains at 224$\times$224 for fast convergence. Stage~2 transitions to 480p, 720p, and 1080p. Stage~3 adds adversarial training. Full details are provided in the supplementary material.

\paragraph{Evaluation Data.}
We evaluate on the UltraVideo~\cite{ultravideo} dataset (\texttt{clips\_short\_1.zip} split), which provides 1,145 high-quality videos at 4K+ resolution covering diverse content from static scenes to fast motion. Each video is first resized to 720$\times$1280 using PIL bicubic interpolation~\cite{umesh2012image, Keys1981CubicCI}, then center-cropped to the target resolution (480p, 720p, or 1080p), and 25-frame clips are extracted. Unless stated otherwise, all evaluations use 25-frame clips. We re-evaluate all baselines using their official implementations and checkpoints on a single H100 GPU.

\paragraph{Evaluation Metrics.}
We report PSNR for pixel-level fidelity, LPIPS~\cite{zhang2018unreasonable} for perceptual quality, and Content-Debiased FVD (rCD-FVD)~\cite{ge2024content} for temporal realism. Standard FVD~\cite{Unterthiner2019FVD} is known to be biased toward per-frame spatial quality rather than temporal consistency; rCD-FVD removes this content bias, providing a more faithful measure of temporal coherence. We refer to rCD-FVD as rFVD for simplicity. Throughput (FPS) is measured as decoded frames per second, and Mem denotes peak GPU memory in GB during decoding. FlashDecoder uses streaming mode while other methods use their native inference modes.

\subsection{Effectiveness of Each Component}
\label{sec:ablation}

Table~\ref{tab:ablation_component} ablates each component using FlashDecoder-S (Table~\ref{tab:ablation_extended}). We evaluate on 480p with 17-frame clips for fast iteration. The \emph{baseline} is a vanilla Transformer with full block causal attention (no windowing). Components are added incrementally: rows (a)--(d) train for 100K iterations at 224$\times$224; rows (e)--(g) add scale-up, streaming training at 480p/720p, and adversarial training, respectively.

\paragraph{Baseline and Streaming Efficiency.}
The baseline's KV cache grows linearly with video length, collapsing throughput from 331.4 to 16.6 FPS. We additionally compare \emph{Sliding-Window Causal Attention} (SW-CA), a separate reference that restricts attention to a fixed window during training. SW-CA restores stable throughput (333.8 FPS) but still requires mask materialization, causing out-of-memory failures at 480p and 720p. Both SW-CA and GQA increase rFVD over the baseline, indicating that limiting temporal context without dedicated temporal modeling hurts temporal coherence.

\paragraph{Architectural Components.}
Among the added components, Temporal Refinement (TR) yields the largest single gain (rFVD: 121.87$\rightarrow$86.94), confirming that raw channel expansion alone produces temporally inconsistent frames that benefit from dedicated refinement. Spatial Upsampling (SU) improves pixel fidelity (PSNR, LPIPS) but slightly increases rFVD, suggesting a trade-off between spatial detail and temporal smoothness. Model scaling provides consistent improvements across all metrics.

\paragraph{Streaming Training and Adversarial Loss.}
The most significant jump occurs when streaming training is introduced (row e$\rightarrow$f). Streaming training enables high-resolution fine-tuning at 480p and 720p, directly improving reconstruction quality by closing the domain gap between 224$\times$224 pretraining and the evaluation resolution. Adversarial training (row f$\rightarrow$g) trades a small PSNR decrease (37.52$\rightarrow$37.08) for sharper outputs (rFVD: 12.29$\rightarrow$10.77), a well-known characteristic of GAN-based losses~\cite{Goodfellow2014GAN}.

\subsection{Video Reconstruction Results}
\label{sec:video_results}

Table~\ref{tab:video_benchmark} compares FlashDecoder against state-of-the-art video decoders on UltraVideo at 480p, 720p, and 1080p, grouped by compression ratio (T$\times$H$\times$W). We train separate FlashDecoder-XL models for each compression group: one on the Wan2.1 encoder latent space (4$\times$8$\times$8) and one on the Wan2.2 encoder latent space (4$\times$16$\times$16).

\paragraph{Convolution-based Models.}
In the 4$\times$16$\times$16 group, FlashDecoder-XL closely matches Wan2.2~\cite{wan2025} in PSNR and LPIPS across all three resolutions while streaming $3.6{\times}$--$4.7{\times}$ faster with up to $11{\times}$ lower peak memory. The memory gap is most pronounced at 1080p: 3.7\,GB versus 41.0\,GB. In the 4$\times$8$\times$8 group, FlashDecoder-XL achieves reconstruction quality comparable to Wan2.1, though it falls behind HunyuanVideo~\cite{kong2024hunyuanvideo}. While rFVD is moderately higher than convolutional baselines in both groups, we note that these are production-level decoders likely trained with significantly more compute and data than our single-node setup. Visualizations are shown in Figure~\ref{fig:visual_comparison} and the supplementary material.

\paragraph{Transformer-based Decoders.}
Existing Transformer decoders face a quality--streaming trade-off: OmniTokenizer~\cite{wang2024omnitokenizer} supports streaming but at low quality, while MAGI-1~\cite{teng2025magi} and AToken~\cite{atoken} achieve better quality through bidirectional attention but cannot stream and degrade in throughput with video length. FlashDecoder-XL resolves this by achieving higher reconstruction quality than both MAGI-1 and AToken while streaming at $2.5{\times}$--$13{\times}$ higher throughput, without padding or blending.

\input{Table_Figure/generalization}
\input{Table_Figure/ablation_wfrm}

\paragraph{Generalization Across Latent Spaces.} FlashDecoder is not tied to a specific encoder. Table~\ref{tab:generalization} shows results when training on the Wan2.1 latent space, which uses $8{\times}$ spatial compression instead of $16{\times}$. Because $8{\times}$ compression produces $2{\times}$ more spatial tokens per dimension ($4{\times}$ total), we apply PixelUnshuffle with factor 2 to the latent, folding the extra spatial dimensions into channels. This yields the same spatial token count $L_{\text{frm}}$ as the $16{\times}$ setting, so the Transformer backbone operates with minimal architecture changes. FlashDecoder-XL achieves comparable quality to Wan2.1 (37.46 vs.\ 37.43 PSNR at 720p) with $4.8{\times}$ higher throughput and $6.8{\times}$ lower memory.

\begin{figure}[t]
  \centering
  \includegraphics[width=1.0\linewidth]{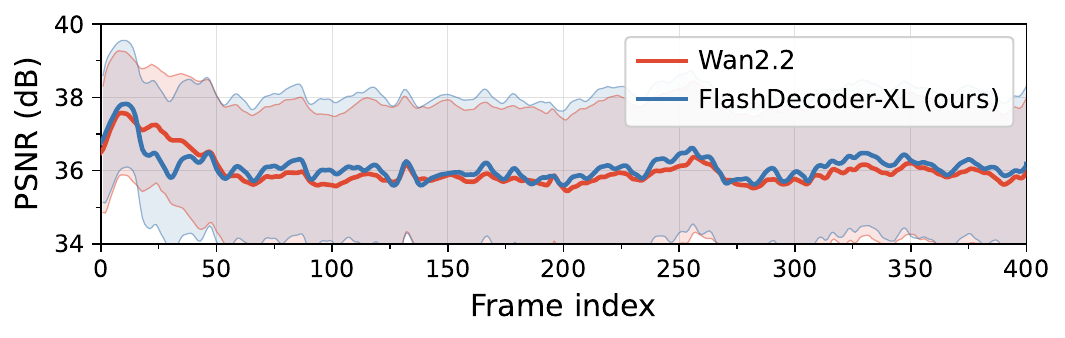}
  \vspace{-2mm}
  \caption{\textbf{Per-frame PSNR on long videos at 720p.} Averaged over 40 videos ($>$400 frames each) from UltraVideo. FlashDecoder maintains stable quality with constant memory regardless of video length.}
  \label{fig:long_video}
\end{figure}

\paragraph{Long Video Decoding.} Because the KV cache window is fixed, FlashDecoder maintains constant memory regardless of video length. We assign RoPE positions relative to the current window rather than the absolute frame index, so positional encodings always stay within the range seen during training, enabling theoretically infinite-length decoding. Figure~\ref{fig:long_video} shows per-frame PSNR on 720p videos exceeding 400 frames. FlashDecoder-XL maintains stable reconstruction quality throughout. Wan2.2 also supports streaming but consumes significantly more memory per frame.

\input{Table_Figure/ablation_scaling_law}

\subsection{Window Size Ablation}
\label{sec:wfrm}
Table~\ref{tab:wfrm} ablates the window size $W_{\text{frm}}$. Performance is stable across $W_{\text{frm}} \in \{2, 3, 4\}$, with $W_{\text{frm}}{=}2$ providing the best trade-off: similar quality to $W_{\text{frm}}{=}4$ with lower memory (2.4 vs.\ 2.6\,GB) and higher throughput (76.3 vs.\ 60.6 FPS). This suggests that for latent decoding, attending to just one previous latent frame provides sufficient temporal context.

\subsection{Scaling Analysis}
\label{sec:scaling}

Table~\ref{tab:ablation_extended} shows four FlashDecoder variants trained at 224$\times$224 for 150K iterations and evaluated at 480p. All variants share the same architecture (GQA Transformer blocks with temporal refinement and PixelShuffle spatial upsampling) and differ in width ($D$), depth, number of heads ($N$), and KV groups ($G$). FlashDecoder-XL, the largest variant, uses $D{=}1536$ with 20 backbone blocks, 2 temporal refinement blocks, 24 attention heads, and 3 KV groups (full configuration in supplementary material). Reconstruction quality improves steadily with model size: PSNR rises from 30.90 to 33.81 and rFVD drops from 89.23 to 31.00. Throughput decreases with scale (254.5 $\rightarrow$ 166.0 FPS), yet even FlashDecoder-XL comfortably maintains real-time streaming performance at all resolutions.

\subsection{Inference Optimization}
\label{sec:optimization}
FlashDecoder's streaming architecture is particularly amenable to inference optimization because each frame-level forward pass has a fixed, data-independent compute graph. We apply four progressive optimizations, each building on the previous: (1)~\texttt{torch.compile} fuses elementwise operations (RMSNorm, SiLU, residual additions) into single GPU kernels, reducing memory bandwidth pressure; (2)~CUDA graph capture eliminates per-step Python dispatch and kernel launch overhead by replaying the entire forward pass as a single graph; (3)~precomputed RoPE tables and a FlashAttention-3 custom operator remove dynamic allocations and graph breaks that would otherwise prevent end-to-end graph capture; (4)~static-calibrated FP8 quantization of all MLP layers exploits H100 FP8 Tensor Cores for higher matmul throughput. The first three optimizations are lossless. FP8 quantization incurs a quality trade-off: PSNR drops by 0.06--0.71\,dB and rFVD increases by up to 0.94 depending on resolution (Table~\ref{tab:video_benchmark}). These optimizations are complementary and stack multiplicatively. The resulting FlashDecoder-XL-Opt (Table~\ref{tab:video_benchmark}) achieves up to $12{\times}$ higher throughput than Wan2.2 at 480p while using under 2\,GB peak memory on the 4$\times$16$\times$16 latent space.

%% file: Table_Figure/ablation.tex
\begin{table}[t]
\centering
\caption{\textbf{Component ablation.}
We incrementally add architectural and training components to a blockwise causal vanilla Transformer decoder.
SW-CA: Sliding-Window Causal Attention;
GQA: Grouped-Query Attention~\cite{ainslie2023gqa};
TR: Temporal Refinement;
SU: Spatial Upsampling;
Scale-up: model scale-up from 56.8M to 769.3M parameters;
Streaming: streaming training with a rolling KV cache;
Adv: adversarial training.
All evaluations are performed on 480p videos with 17 frames for efficient ablation.
}
\label{tab:ablation_component}
\scalebox{0.85}{%
\begin{tabular}{lcccc}
\toprule
\textbf{Components} & \textbf{PSNR}$\uparrow$ & \textbf{LPIPS}$\downarrow$ & \textbf{rFVD}$\downarrow$ & \textbf{FPS}$\uparrow$ \\
\midrule
Baseline & 30.30 & 0.16 & 117.77 & 331.4$\rightarrow$16.6 \\
(a)\;+ SW-CA  & 30.20 & 0.15 & 136.08 & 333.8 \\
(b)\;+ GQA & 30.13 & 0.14 & 121.87 & 340.7 \\
(c)\;+ TR & 31.05 & 0.13 & 86.94 & 260.3 \\
(d)\;+ SU & 31.49 & 0.12 & 96.19 & 262.1 \\
\midrule
(e)\;+ Scale-up  & 32.56 & 0.09 & 44.74 & 166.0 \\
(f)\;\quad+ Streaming         & 37.52 & 0.05 & 12.29 & 166.0 \\
(g)\;\quad\quad+ Adv               & 37.08 & 0.05 & 10.77 & 166.0 \\
\bottomrule
\end{tabular}
}
\end{table}

%% file: Table_Figure/benchmark_video.tex
\begin{table*}[t!]
\centering
\caption{\textbf{Video reconstruction benchmark.}
Results on UltraVideo~\cite{ultravideo} at 480p, 720p, and 1080p. We report PSNR, LPIPS~\cite{zhang2018unreasonable}, rFVD~\cite{Unterthiner2019FVD, ge2024content}, throughput (FPS), and peak GPU memory (Mem, GB). All measurements use 25-frame clips on a single H100 GPU. FlashDecoder uses streaming mode; other methods use their native inference modes. $^{\dagger}$Causal/streaming. $^{*}$256px only.}
\small
\resizebox{1.0\textwidth}{!}{
\setlength{\tabcolsep}{3pt}
\renewcommand{\arraystretch}{1.05}
\begin{tabular}{@{}lcccccccccccccccccc@{}}
\toprule
\multirow{2}{*}[-0.5ex]{\textbf{Method}} & \multirow{2}{*}[-0.5ex]{\makecell{\textbf{Params}\\\textbf{(M)}}} &
\multicolumn{5}{c}{\textbf{480p}} & & \multicolumn{5}{c}{\textbf{720p}} & & \multicolumn{5}{c}{\textbf{1080p}} \\
\cmidrule(lr){3-7} \cmidrule(lr){9-13} \cmidrule(lr){15-19}
 &  & \textbf{PSNR} & \textbf{LPIPS} & \textbf{rFVD} & \textbf{FPS} & \textbf{Mem} & & \textbf{PSNR} & \textbf{LPIPS} & \textbf{rFVD} & \textbf{FPS} & \textbf{Mem} & & \textbf{PSNR} & \textbf{LPIPS} & \textbf{rFVD} & \textbf{FPS} & \textbf{Mem} \\
\midrule
\rowcolor{gray!20}
\multicolumn{19}{l}{\textit{4$\times$8$\times$8 Compression}} \\
\addlinespace[1pt]
OmniTokenizer$^{*\dagger}$~\cite{wang2024omnitokenizer} & 26.9 & 17.25 & 0.20 & 60.40 & 2333.6 & 0.3 & & -- & -- & -- & -- & -- & & -- & -- & -- & -- & -- \\
HunyuanVideo$^{\dagger}$~\cite{kong2024hunyuanvideo} & 146.1 & 37.84 & 0.05 & 7.67 & 11.1 & 7.2 & & 38.65 & 0.05 & 10.25 & 4.8 & 7.3 & & 41.93 & 0.05 & 7.74 & 2.1 & 7.6 \\
Wan2.1$^{\dagger}$~\cite{wan2025} & 73.3 & 36.63 & 0.04 & 9.91 & 36.0 & 7.3 & & 37.43 & 0.04 & 12.43 & 15.9 & 16.4 & & 40.36 & 0.05 & 9.94 & 6.9 & 36.8 \\
Wan2.1-TAEHV$^{\dagger}$~\cite{BoerBohan2025TAEHV} & 9.8 & 32.65 & 0.11 & 33.37 & 663.3 & 5.7 & & 33.43 & 0.11 & 35.52 & 309.8 & 13.1 & & 36.80 & 0.11 & 33.37 & 135.1 & 23.7 \\
MAGI-1~\cite{teng2025magi} & 306.5 & 35.08 & 0.14 & 42.24 & 43.8 & 2.3 & & 34.82 & 0.15 & 46.23 & 9.3 & 3.8 & & 37.02 & 0.17 & 60.52 & 2.0 & 7.1 \\
\textbf{FlashDecoder-XL$^{\dagger}$} & 750.4 & 35.92 & 0.05 & 11.62 & 162.3 & 1.8 & & 37.46 & 0.05 & 12.13 & 76.1 & 2.4 & & 40.74 & 0.05 & 13.34 & 25.9 & 3.6 \\
\textbf{FlashDecoder-XL-Opt$^{\dagger}$} & 750.4 & 35.72 & 0.05 & 12.42 & 449.0 & 1.2 & & 37.17 & 0.05 & 14.02 & 152.0 & 1.5 & & 40.44 & 0.05 & 14.12 & 43.0 & 2.3 \\
\midrule
\rowcolor{gray!20}
\multicolumn{19}{l}{\textit{4$\times$16$\times$16 Compression}} \\
\addlinespace[1pt]
Wan2.2$^{\dagger}$~\cite{wan2025} & 555.0 & 37.47 & 0.04 & 7.97 & 36.6 & 9.6 & & 38.29 & 0.04 & 10.39 & 16.1 & 19.3 & & 41.49 & 0.04 & 8.16 & 7.1 & 41.0 \\
Wan2.2-TAEHV$^{\dagger}$~\cite{BoerBohan2025TAEHV} & 9.9 & 30.16 & 0.21 & 88.54 & 2329.5 & 1.4 & & 30.68 & 0.21 & 88.32 & 1177.6 & 3.3 & & 33.55 & 0.21 & 88.49 & 524.9 & 7.4 \\
AToken~\cite{atoken} & 415.1 & 35.71 & 0.08 & 19.44 & 65.3 & 0.7 & & 36.72 & 0.09 & 22.91 & 28.1 & 1.5 & & 40.18 & 0.09 & 25.67 & 10.1 & 3.3 \\
\textbf{FlashDecoder-XL$^{\dagger}$} & 769.3 & 37.08 & 0.05 & 10.77 & 166.0 & 1.9 & & 38.38 & 0.05 & 12.75 & 76.3 & 2.4 & & 41.55 & 0.05 & 12.08 & 25.4 & 3.7 \\
\textbf{FlashDecoder-XL-Opt$^{\dagger}$} & 769.3 & 37.02 & 0.05 & 11.22 & 441.0 & 1.1 & & 37.85 & 0.05 & 12.22 & 151.0 & 1.3 & & 40.84 & 0.05 & 13.02 & 43.0 & 1.6 \\
\bottomrule
\end{tabular}
}
\vspace{-1mm}
\label{tab:video_benchmark}
\end{table*}

%% file: Table_Figure/generalization.tex
\begin{table}[t!]
\centering
\caption{\textbf{Generalization across VAE latent spaces.} FlashDecoder-XL trained on different encoder latent spaces, evaluated at 720p with 25 frames. Mem denotes peak GPU memory in GB. FlashDecoder generalizes across latent spaces with comparable quality while achieving ${\sim}5{\times}$ higher throughput and up to $8{\times}$ lower peak memory.}
\label{tab:generalization}
\small
\resizebox{\columnwidth}{!}{%
\setlength{\tabcolsep}{3pt}
\begin{tabular}{llccccc}
\toprule
Encoder & Decoder & PSNR$\uparrow$ & LPIPS$\downarrow$ & rFVD$\downarrow$ & FPS$\uparrow$ & Mem$\downarrow$ \\
\midrule
Wan2.1 & Wan2.1 & 37.43 & 0.04 & 12.43 & 15.9 & 16.4 \\
Wan2.1 & FlashDecoder-XL & 37.46 & 0.05 & 12.13 & 76.1 & 2.4 \\
\midrule
Wan2.2 & Wan2.2 & 38.29 & 0.04 & 10.39 & 16.1 & 19.3 \\
Wan2.2 & FlashDecoder-XL & 38.38 & 0.05 & 12.75 & 76.3 & 2.4 \\
\bottomrule
\end{tabular}}
\end{table}

%% file: Table_Figure/ablation_wfrm.tex
\begin{table}[t!]
\centering
\caption{\textbf{Effect of window size $W_{\text{frm}}$.} Evaluated at 720p with 25 frames using FlashDecoder-XL. Mem denotes peak GPU memory in GB. Performance is stable across window sizes; $W_{\text{frm}}{=}2$ provides a good trade-off between quality and memory.}
\label{tab:wfrm}
\small
\setlength{\tabcolsep}{3pt}
\begin{tabular}{cccccc}
\toprule
$W_{\text{frm}}$ & PSNR$\uparrow$ & LPIPS$\downarrow$ & rFVD$\downarrow$ & FPS$\uparrow$ & Mem$\downarrow$ \\
\midrule
2 & 38.38 & 0.05 & 12.75 & 76.3 & 2.4 \\
3 & 38.13 & 0.05 & 12.71 & 67.8 & 2.5 \\
4 & 38.49 & 0.05 & 12.87 & 60.6 & 2.6 \\
\bottomrule
\end{tabular}
\end{table}

%% file: Table_Figure/ablation_scaling_law.tex
\begin{table*}[t!]
\centering
\caption{\textbf{Model scaling.} FlashDecoder variants trained for 150K iterations at 224$\times$224 (Stage~1 only) and evaluated on UltraVideo at 480p with 17 frames for fast iteration. Numbers are not directly comparable to Table~\ref{tab:video_benchmark}, which uses the full three-stage training. Mem denotes peak GPU memory in GB.}
\small
\setlength{\tabcolsep}{3pt}
\begin{tabular}{lccccccccc}
\toprule
\textbf{Model} & \textbf{Depth} & \textbf{Width ($D$)} & \textbf{Heads ($N$)} & \textbf{KV Groups ($G$)} & \makecell{\textbf{Params}\\(M)} & \textbf{PSNR}$\uparrow$ & \textbf{rFVD}$\downarrow$ & \textbf{FPS}$\uparrow$ & \textbf{Mem}$\downarrow$ \\
\midrule
FlashDecoder-S  & 12 &  512 &  8 & 2 &  56.8  & 30.90 & 89.23 & 254.5 & 0.3 \\
FlashDecoder-B  & 16 &  768 & 12 & 3 & 161.7  & 31.15 & 72.36 & 205.2 & 0.6 \\
FlashDecoder-L  & 20 & 1024 & 16 & 4 & 348.0  & 32.13 & 63.81 & 164.3 & 1.0 \\
FlashDecoder-XL & 20 & 1536 & 24 & 3 & 769.3  & 33.81 & 31.00 & 166.0 & 1.9 \\
\bottomrule
\end{tabular}
\label{tab:ablation_extended}
\end{table*}

%% file: Sections/5_conclusion.tex
\section{Conclusion}
We introduced FlashDecoder, a pure-Transformer latent-to-pixel decoder that achieves real-time streaming by processing one frame at a time with a fixed-size rolling KV cache. Two findings stand out from our experiments. First, enforcing causality through temporal processing order rather than explicit masks removes the memory barrier to high-resolution training, enabling stable training up to 1080p. Second, this high-resolution training closes the reconstruction quality gap with convolutional decoders, a gap that has limited prior Transformer decoders. On both the Wan2.1 and Wan2.2 latent spaces, FlashDecoder-XL matches convolutional decoder reconstruction quality while delivering $3.6{\times}$--$4.7{\times}$ faster streaming throughput and up to $11{\times}$ lower GPU memory. With architecture-aware inference optimizations, the throughput gap widens to $12{\times}$, enabling real-time high-resolution decoding on a single GPU. A natural next step is to pair FlashDecoder with a streaming Transformer encoder and train the full VAE from scratch, potentially unlocking latent spaces better suited to Transformer-based generation and decoding.

\clearpage
\section*{Acknowledgments}
Minguk Kang is a participating researcher at POSTECH and a full-time employee at Pika Labs.
We thank Joonghyuk Shin for helpful discussions and assistance with MotionStream experiments, and Zhicheng Sun and Cade Li for valuable discussions.
This work was supported by Samsung Electronics Co., Ltd.\ (Samsung AI Center) and the IITP grants (RS-2022-II220290, RS-2022-II220926, RS-2019-II191906) funded by the Korea government (MSIT).

%% file: Sections/7_appendix.tex
This supplementary material provides training specifications (Section~\ref{sec:supp_training}), inference protocols (Section~\ref{sec:inference}), additional visual results (Section~\ref{sec:qualitative}), and limitations and future directions (Section~\ref{sec:limitations}).

\section{Training Specifications}
\label{sec:supp_training}

\subsection{Dataset Details}

\noindent\textbf{Image Data.}
We utilize DataComp-small~\cite{gadre2023datacomp}, comprising 12.8M image-text pairs. During preprocessing, we apply probabilistic augmentation that randomly selects among random cropping (40\%), center cropping (30\%), or resizing (30\%) to the target resolution. Images smaller than the target resolution are filtered to prevent upsampling artifacts.

\noindent\textbf{Video Data.}
Our video corpus combines Kinetics-600~\cite{carreira2018short} and an internal high-resolution collection of approximately 200K clips. From each video, we sample 17 consecutive frames at native frame rate. Preprocessing employs a two-stage spatial transformation: frames are first resized so that the shorter side matches the target resolution (480p, 720p, or 1080p depending on the training stage) while preserving the original aspect ratio, then cropped to the target resolution using either center crop (60\%) or random crop (40\%). Videos below the target resolution are filtered out. All resizing uses anti-aliased PIL bicubic interpolation~\cite{umesh2012image, Keys1981CubicCI}.

\subsection{Multi-Stage Training Protocol}

FlashDecoder follows a sequential three-stage training protocol, with each stage building upon the previous one. Training hyperparameters are summarized in Table~\ref{tab:train_hparams}.

\noindent\textbf{Stage 1: Low-Resolution Pre-training.}
This stage establishes fundamental reconstruction capabilities at reduced computational cost. We train on 224$\times$224$\times$17 video clips and 256$\times$256 images with a 2:8 image-to-video sampling ratio to balance temporal coherence with spatial fidelity. Training proceeds for 200K iterations with batch size 16. The reconstruction objective combines L1 loss and perceptual loss~\cite{zhang2018unreasonable} with weights of 1.0 and 0.1, respectively.

\noindent\textbf{Stage 2: High-Resolution Training.}
We transition to higher resolutions to minimize the domain gap between training and inference. The model is trained on 480p clips (480$\times$832$\times$17), 720p clips (720$\times$1280$\times$17), 1080p clips (1080$\times$1920$\times$17), and 512$\times$512 images. This diverse resolution mixture enables the model to handle varying spatial resolutions during inference. We reduce the learning rate by 10$\times$ and train for 100K iterations with batch size 8. Loss weights are adjusted to 1.0 for L1 and 0.25 for perceptual loss. The perceptual loss is computed on random 224$\times$224 crops to keep memory manageable at high resolutions.

\noindent\textbf{Stage 3: Adversarial Post-training.}
To enhance fine-grained details, we introduce adversarial training using the same data configuration as Stage 2 but excluding 1080p clips due to the additional memory overhead of the discriminator. This stage enables the decoder to synthesize sharper high-frequency textures that reconstruction losses alone cannot capture. We extend VQGAN's 2D PatchGAN discriminator~\cite{vqgan2021repo, isola2017image} to 3D for spatiotemporal processing, and train it with non-saturating logistic loss~\cite{Goodfellow2014GAN} and R1 regularization~\cite{Mescheder2018ICML}. Both the perceptual and adversarial losses are computed on random 224$\times$224 crops of the decoded output.

\section{Inference Protocols}
\label{sec:inference}

For fair comparison, we re-evaluate all baseline models using their official repository implementations and released checkpoints under identical settings on a single NVIDIA H100 GPU (80GB).

\subsection{Throughput Measurement}

To ensure fair comparison, we evaluate each model in its officially supported inference mode. Wan2.2-TAEHV~\cite{BoerBohan2025TAEHV}, Wan2.2~\cite{wan2025}, and FlashDecoder natively support streaming and are evaluated accordingly. Other VAEs (HunyuanVideo, AToken, MAGI-1) process entire clips in batch mode using their official implementations, as forcing them into a streaming setup would require chunking and blending that degrades their reconstruction quality. Throughput is measured in frames per second (FPS), calculated as total decoded frames divided by total decoding time.

\begin{table*}[t!]
\centering
\caption{{\bf Hyperparameters for FlashDecoder-XL training on the Wan2.2 latent space.} We report the training configurations for each stage. Stage 1 focuses on low-resolution pre-training, Stage 2 transitions to high-resolution training, and Stage 3 introduces adversarial post-training. For additional technical details, please refer to the original papers: LPIPS~\cite{zhang2018unreasonable}, R1 regularization~\cite{Mescheder2018ICML}, AdamW optimizer~\cite{loshchilov2018decoupled}, and 3D PatchGAN~\cite{esser2021taming, isola2017image}. DDP denotes Distributed Data Parallel.}
\resizebox{0.95\textwidth}{!}
{
\begin{tabular}{lccc}
\cmidrule[1.0pt]{1-4}
Hyperparameters & Stage 1 & Stage 2 & Stage 3 \\
\cmidrule[1.0pt]{1-4}
\multicolumn{4}{l}{\textit{Model Architecture}} \\
\cmidrule[0.5pt]{1-4}

Latent channels ($C'$) & 48 & 48 & 48 \\
Model dimension ($D$) & 1536 & 1536 & 1536 \\
$\#$ of Transformer blocks & 20 & 20 & 20 \\
$\#$ of Temporal refinement Transformer blocks & 2 & 2 & 2 \\
Attention heads ($N$) & 24 & 24 & 24 \\
KV groups ($G$) & 3 & 3 & 3 \\
MLP expansion & 4.0 & 4.0 & 4.0 \\
Temporal compression ($r_{\text{t}}$) & 4 & 4 & 4 \\
Spatial compression ($r_{\text{s}}$) & 16 & 16 & 16 \\
Window size ($W_{\text{frm}}$)& 2 & 2 & 2 \\
\cmidrule[0.5pt]{1-4}
\multicolumn{4}{l}{\textit{Data Configuration}} \\
\cmidrule[0.5pt]{1-4}
Video resolution & 224$\times$224$\times$17 & 480p / 720p / 1080p $\times$17 & 480p / 720p $\times$17 \\
Image resolution & 256$\times$256 & 512$\times$512 & 512$\times$512 \\
Sampling ratio & 8:2 (video:image) & 2:4:2:2 (480p:720p:1080p:image) & 2:6:2 (480p:720p:image) \\
\cmidrule[0.5pt]{1-4}
\multicolumn{4}{l}{\textit{Loss Configuration}} \\
\cmidrule[0.5pt]{1-4}
L1 loss weight ($\lambda_{\text{L1}}$) & 1.0 & 1.0 & 1.0 \\
Perceptual loss weight ($\lambda_{\text{LPIPS}}$) & 0.1 & 0.25 & 0.25 \\
Adversarial loss type & - & - & Logistic \\
Adversarial loss weight ($\lambda_{\text{adv}}$) & - & - & 1e-4 \\
R1 regularization weight & - & - & 0.1024 \\
R1 interval & - & - & 16 \\
\cmidrule[0.5pt]{1-4}
\multicolumn{4}{l}{\textit{Decoder $\&$ Decoder-related Optimization}} \\
\cmidrule[0.5pt]{1-4}
Optimizer & AdamW & AdamW & AdamW \\
Batch size & 16 & 8 & 8 \\
Learning rate & 1e-4 & 1e-5 & 1e-5 \\
AdamW $\beta_{1}$ & 0.9 & 0.9 & 0.9 \\
AdamW $\beta_{2}$ & 0.999 & 0.999 & 0.999 \\
Weight decay & 0.01 & 0.01 & 0.01 \\
EMA & - & - & 0.9999 \\
EMA warmup step & - & - & 2000 \\
Precision & bfloat16 & bfloat16 & bfloat16 \\
\cmidrule[0.5pt]{1-4}
\multicolumn{4}{l}{\textit{Discriminator $\&$ Discriminator-related Optimization}} \\
\cmidrule[0.5pt]{1-4}
Architecture & - & - & 3D PatchGAN \\
$\#$ of conv layers & - & - & 5 \\
Base channels & - & - & 128 \\
Learning rate & - & - & 1e-5 \\
AdamW $\beta_{1}$ & - & - & 0.0 \\
AdamW $\beta_{2}$ & - & - & 0.9 \\
Weight decay & - & - & 0.01 \\
Precision & bfloat16 & bfloat16 & bfloat16 \\
\cmidrule[0.5pt]{1-4}
\multicolumn{4}{l}{\textit{Training specifications}} \\
\cmidrule[0.5pt]{1-4}
Training iterations & 200K & 100K & 20K \\
Distributed training & DDP & DDP & DDP \\
GPU type & H100 80GB & H100 80GB & H100 80GB \\
$\#$ GPUs & 8 & 8 & 8 \\
\cmidrule[1.0pt]{1-4}
\end{tabular}}
\label{tab:train_hparams}
\end{table*}

\section{More Visual Results}
\label{sec:qualitative}
\begin{figure*}[t!]
  \centering
  \includegraphics[width=1.0\linewidth]{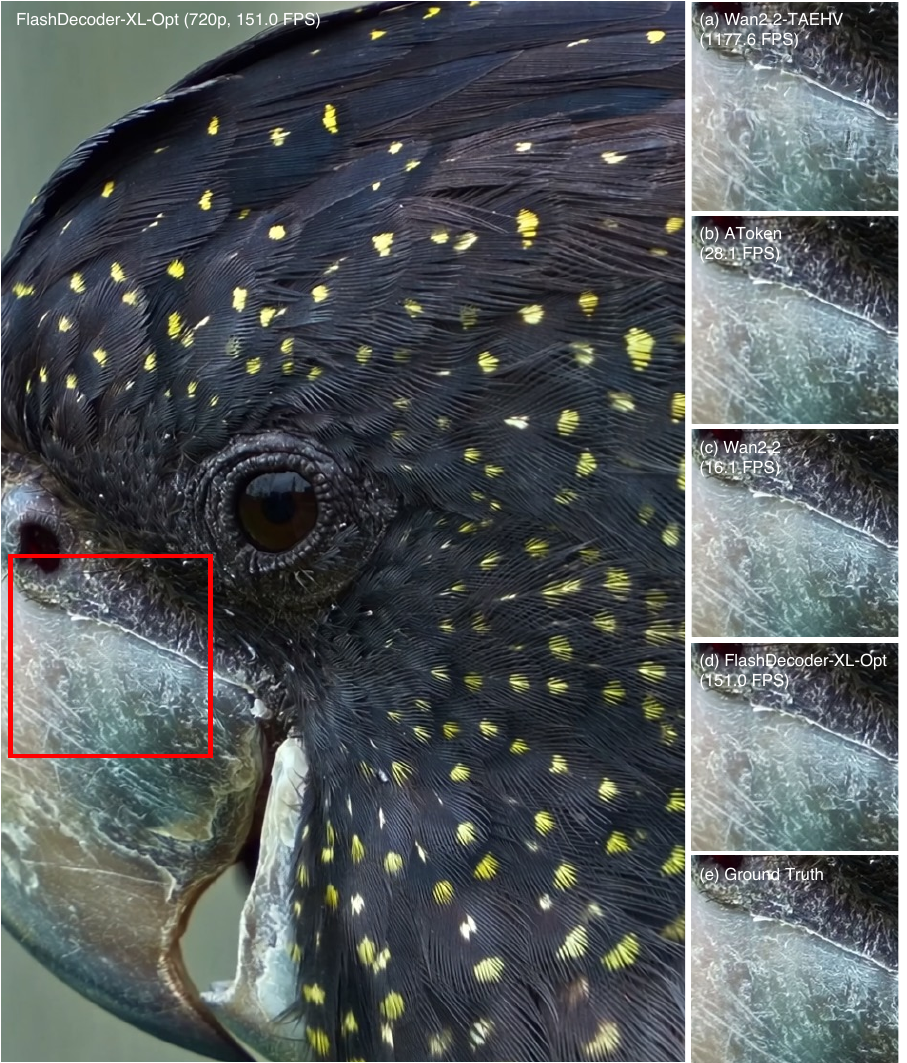}
\caption{\textbf{Qualitative comparison of 720p reconstruction results.}
We compare reconstructed frames from video decoders with 4$\times$ temporal and 16$\times$ spatial compression: (a) Wan2.2-TAEHV~\cite{BoerBohan2025TAEHV}, (b) AToken~\cite{atoken}, (c) Wan2.2~\cite{wan2025}, (d) our FlashDecoder-XL-Opt, and (e) ground truth.
(a) and (b) produce blurry reconstructions, while (c) and (d) yield visually comparable outputs, yet (d) achieves over $9{\times}$ higher throughput.}
  \label{fig:visual_comparison2}
\end{figure*}

\begin{figure*}[t!]
  \centering
  \includegraphics[width=1.0\linewidth]{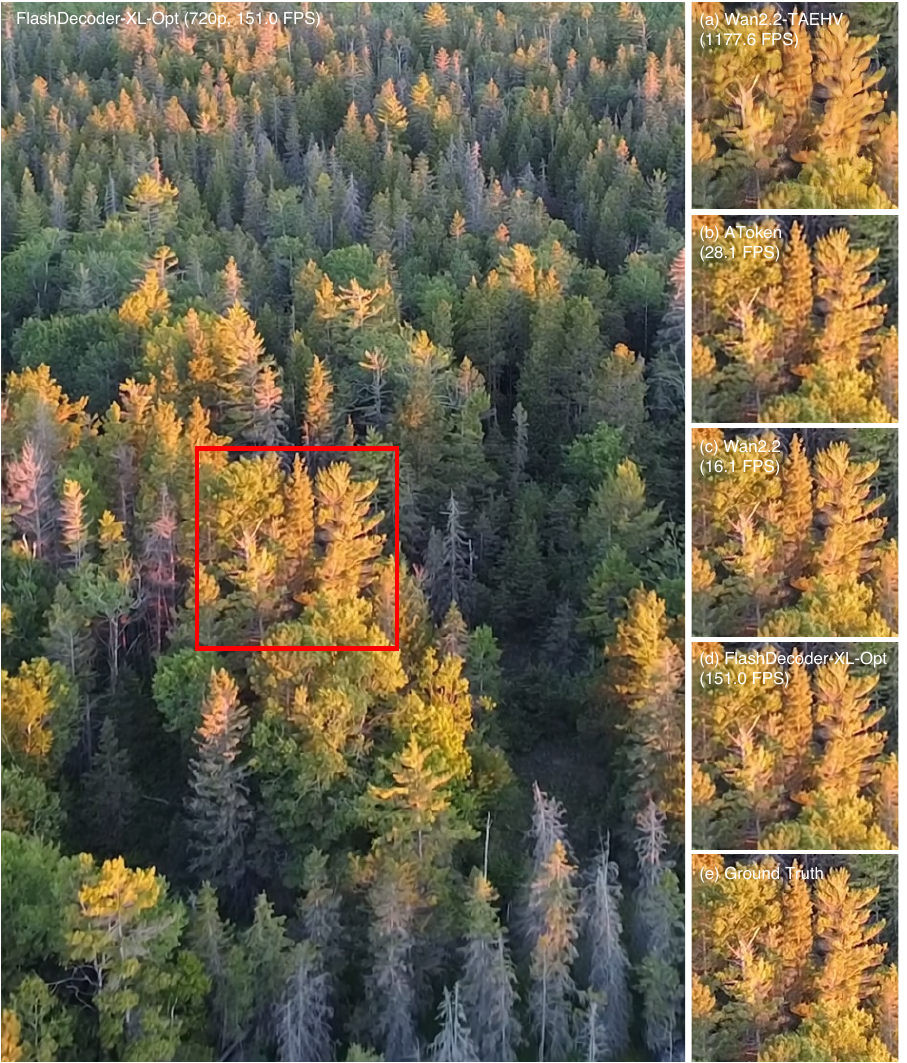}
\caption{\textbf{Qualitative comparison of 720p reconstruction results.}
We compare reconstructed frames from video decoders with 4$\times$ temporal and 16$\times$ spatial compression: (a) Wan2.2-TAEHV~\cite{BoerBohan2025TAEHV}, (b) AToken~\cite{atoken}, (c) Wan2.2~\cite{wan2025}, (d) our FlashDecoder-XL-Opt, and (e) ground truth.
(a) fails to decode fine details such as tree branches and foliage, while (b) produces blurry reconstructions. (c) and (d) yield sharper results, yet (d) achieves over $9{\times}$ higher throughput.}
  \label{fig:visual_comparison3}
\end{figure*}

\begin{figure*}[t!]
  \centering
  \includegraphics[width=1.0\linewidth]{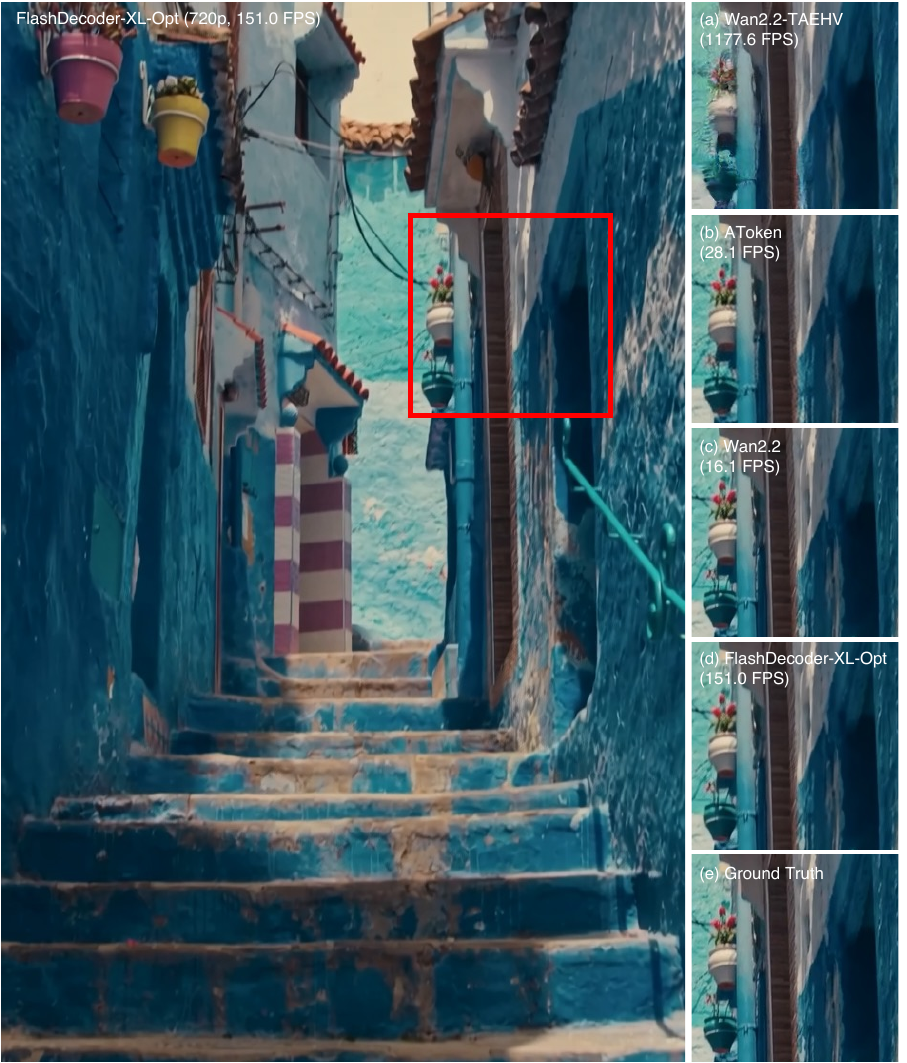}
\caption{\textbf{Qualitative comparison of 720p reconstruction results.}
We compare reconstructed frames from video decoders with 4$\times$ temporal and 16$\times$ spatial compression: (a) Wan2.2-TAEHV~\cite{BoerBohan2025TAEHV}, (b) AToken~\cite{atoken}, (c) Wan2.2~\cite{wan2025}, (d) our FlashDecoder-XL-Opt, and (e) ground truth.
(a) struggles to decode wall textures near the flowerpot, while (b) produces blurry details in the flowerpot region. (c) and (d) yield visually comparable outputs, with (c) appearing to synthesize marginally finer details, particularly around the flower petals. (d) achieves over $9{\times}$ higher throughput.}
  \label{fig:visual_comparison4}
\end{figure*}

Figures~\ref{fig:visual_comparison2}--\ref{fig:visual_comparison4} show additional 720p reconstruction results from Wan2.2-TAEHV~\cite{BoerBohan2025TAEHV}, AToken~\cite{atoken}, Wan2.2~\cite{wan2025}, and FlashDecoder-XL-Opt. Wan2.2-TAEHV struggles with fine details; AToken produces smoother but blurrier outputs. Both Wan2.2 and FlashDecoder-XL-Opt produce sharp results, with Wan2.2 showing slightly finer details in some cases. FlashDecoder-XL-Opt achieves over $9{\times}$ higher throughput (151.0 vs.\ 16.1 FPS).

\section{Limitations and Future Directions}
\label{sec:limitations}

\noindent\textbf{Decoder-only design.}
FlashDecoder replaces only the decoder while keeping the pretrained convolutional encoder fixed. The latent space therefore inherits the encoder's characteristics, which favor spatial locality by design; what properties a Transformer-based encoder--decoder pair would learn remains unexplored. Designing a streaming Transformer encoder to pair with FlashDecoder and training the full VAE from scratch is a clear next step. Such an encoder--decoder pair would also eliminate the resolution gap between low-resolution VAE training ($256{\times}256$) and high-resolution diffusion training (720p, 1080p), since our streaming architecture scales to high resolutions. This is particularly relevant for recent end-to-end frameworks such as Unified Latents~\cite{heek2026unified}, which jointly train the VAE and diffusion model. Existing convolutional video VAEs are difficult to jointly train with a diffusion model at 480p or 720p due to their high memory consumption. FlashDecoder's low memory footprint makes such joint training feasible. Training the VAE and diffusion model at the same resolution used during inference would ensure that the latent space has good diffusibility at that resolution, avoiding potential mismatches caused by training at a lower resolution.

\noindent\textbf{rFVD gap.}
FlashDecoder-XL falls short of Wan2.2~\cite{wan2025} and HunyuanVideo~\cite{kong2024hunyuanvideo} in rFVD~\cite{Unterthiner2019FVD, ge2024content}, despite comparable PSNR and LPIPS. We trained on a single 8-GPU node, whereas these production decoders likely used significantly more compute and data. Scaling up model capacity and adversarial training duration is expected to close this gap.

\noindent\textbf{Integration with Representation Autoencoders.}
Recent work on Representation Autoencoders (RAE~\cite{zheng2025diffusion}) pairs frozen pretrained encoders (e.g., DINOv2~\cite{oquab2023dinov2}, SigLIP~\cite{tschannen2025siglip}) with Transformer decoders for image generation. Extending this paradigm to video with a streaming-capable decoder like FlashDecoder is a promising direction.

\clearpage

